\documentclass{article}
\usepackage{spconf,amsmath,graphicx}
\usepackage{amssymb}
\usepackage{epstopdf}
\usepackage{hyperref}
\usepackage{subfig}

\newcommand{\ie}{\textit{i}.\textit{e}.}

\usepackage{xspace}

\usepackage{stfloats}   

\newcommand{\method}{{SDFlow}\xspace}

\usepackage{multirow}
\usepackage{cuted}

\newcommand{\vct}[1]{\boldsymbol{#1}} 
\newcommand{\mat}[1]{\boldsymbol{#1}} 

\newcommand{\app}{\raise.17ex\hbox{$\scriptstyle\sim$}}

\newlength\savewidth

\usepackage{subfig}
\usepackage{overpic}

\hypersetup{
    colorlinks=true,    
    linkcolor=red,
    urlcolor=magenta
}

\title{
	Semantic Latent Decomposition with Normalizing Flows \\ for Face Editing
}

\name{Binglei Li$^{1}$ \qquad Zhizhong Huang$^{1}$ \qquad Hongming Shan$^{2}$ \qquad Junping Zhang$^{1}$$^{\dagger}$\thanks{$\dagger$ Corresponding author}}
\address{$^{1}$Shanghai Key Lab of Intelligent Information Processing, School of Computer Science\\
$^{2}$Institute of Science and Technology for Brain-inspired Intelligence \\Fudan University, Shanghai 200433, China
}
\begin{document}
\topmargin=0mm
%
\maketitle

\begin{abstract}

Navigating in the latent space of StyleGAN has shown effectiveness for face editing. 
However, the resulting methods usually encounter challenges in complicated navigation due to the entanglement among different attributes in the latent space.
To address this issue, this paper proposes a novel framework, termed \method, with a semantic decomposition in original latent space using continuous conditional normalizing flows. 
Specifically, \method decomposes the original latent code into different irrelevant variables by jointly optimizing two components: 
(\textbf{i}) a semantic encoder to estimate semantic variables from input faces and 
(\textbf{ii}) a flow-based transformation module to map the latent code into a semantic-irrelevant variable in Gaussian distribution, conditioned on the learned semantic variables.
To eliminate the entanglement between variables, we employ a disentangled learning strategy under a mutual information framework, thereby providing precise manipulation controls.
Experimental results demonstrate that~\method outperforms existing state-of-the-art face editing methods both qualitatively and quantitatively. The source code is made available at \url{https://github.com/phil329/SDFlow}.

\end{abstract}
\begin{keywords}
Face Editing, Disentangle Learning, Generative Adversarial Network
\end{keywords}

\section{Introduction}
\label{sec:intro}

Face editing is to change the desired facial attributes while keeping image quality and other undesired attributes.
In recent years, the generator of StyleGAN~\cite{karras2019stylegan,karras2020stylegan2} has achieved significant progress in synthesizing high-fidelity images, which presents a semantically abundant latent space. Most importantly, the faces can be manipulated by traversing on such latent space, which reduces the pain in training a good GAN~\cite{goodfellow2014generative} from scratch. In detail, the faces need to be inverted into the latent space of StyleGAN to obtain the corresponding latent code~\cite{richardson2021psp,tov2021e4e}, which can be manipulated by different methods~\cite{abdal2022clip2stylegan,abdal2021styleflow,shen2020interface,yao2021latentTransformer,yang2021discovering,wu2021stylespace,yuan2023dofam,huang2023adaptive}, and then decoded by the pre-trained generator to produce edited faces.

Current methods for face editing can be roughly categorized as supervised and unsupervised.
Unsupervised methods employ principal components to identify editing directions~\cite{harkonen2020ganspace,shen2021closed} or operate under the prior of CLIP~\cite{abdal2022clip2stylegan,radford2021clip,patashnik2021styleclip}. Although these approaches offer meaningful transformations, they still fall short of producing precise user-desired editing without the aid of human annotations.
Supervised methods~\cite{abdal2021styleflow,shen2020interface,yao2021latentTransformer} typically use the attributes-labeled images to identify how to manipulate the faces in latent space. Some of them~\cite{shen2020interface,yao2021latentTransformer} assume that face editing can be achieved by linearly interpolating along certain directions. InterFaceGAN~\cite{shen2020interface} learns the editing directions by training a hyperplane in the latent space to separate the examples with binary attributes. LatentTransformer~\cite{yao2021latentTransformer} trains a transformation network to produce dynamic directions. However, these methods may fail to handle the scenario when linear assumption does not hold.
Alternatively, nonlinear methods~\cite{abdal2021styleflow,huang2023adaptive} aim to learn a nonlinear transformation.
StyleFlow~\cite{abdal2021styleflow} leverages normalizing flows to re-sample the edited latent codes and AdaTrans~\cite{huang2023adaptive} splits the whole transformation into several finer steps. Unfortunately, they have not explicitly disentangled different facial attributes due to the use of binary attributes.

To address these issues, this paper proposes \method to achieve a semantic decomposition of the latent space of StyleGAN using continuous conditional normalizing flows. Specifically, there are two key components in \method: (\textbf{i}) a semantic encoder and (\textbf{ii}) a flow-based transformation module. The semantic encoder produces semantic variables for different attributes directly from input faces. Under the conditions of semantic variables, the flow-based transformation module maps the latent codes into semantic-irrelevant variables. To achieve the disentanglement between variables, we jointly optimize these two components under a mutual information framework~\cite{chen2016infogan}. Moreover, a pre-trained attribute classifier is distilled to inject the supervision signals of human annotations into the semantic variables. Consequently, the edited latent codes can be obtained from the flows by only changing the semantic variables.

Our contributions are summarized as follows: (\textbf{i}) we introduce \method, a novel disentangled non-linear latent navigation framework for face editing. It performs semantic decomposition in the latent space of StyleGAN with a disentangled learning strategy, which thus can eliminate the entanglement between attributes and enhance editing controls.
(\textbf{ii}) Both qualitative and quantitative experiments demonstrate the effectiveness of the proposed method in terms of image quality, editing accuracy, and identity/attribute preservation.

\section{Methodology}

\begin{figure*}[ht]
	\centering
	\includegraphics[width=\linewidth]{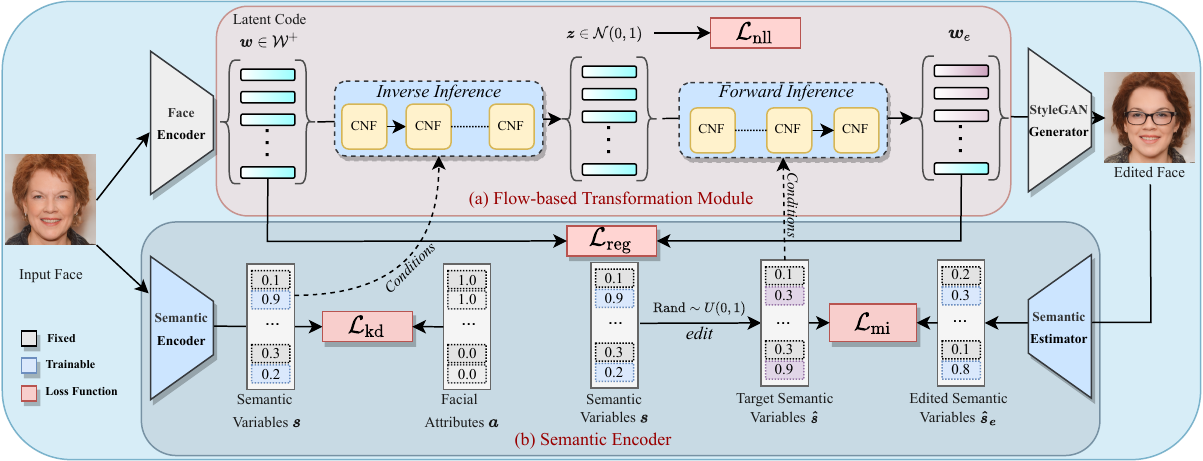}
        \vspace{-20pt}
	\caption{
        Framework of \method.
        \method consists of two components: (a) the flow-based transformation module transforms the original latent codes to semantic-irrelevant variable $\vct{z}$ in Gaussian distribution, conditioned on the semantic variables. The edited faces can be obtained in the forward inference of flows. (b) The semantic encoder estimates the semantic variables of input faces for the flow-based transformation module. They are jointly optimized to achieve a disentangled face editing.
        }
    \label{fig:framework}
\end{figure*}

Fig.~\ref{fig:framework} illustrates the framework of \method. The face image $\mat{I} \in \mathbb{R}^{3\times H\times W}$ are inverted into the latent space $\in \mathcal{W}^+$ of StyleGAN to obtain the layer-wise latent code $\vct{w}=E_{\mathrm{img}}(\mat{I})\in \mathbb{R}^{18\times 512}$, through the pre-trained encoder $E_{\mathrm{img}}$~\cite{tov2021e4e}. Face editing should transform $\vct{w}$ to new latent codes $\vct{w}_e$ for manipulating the faces into $G(\vct{w}_e)$ with target attributes, using the pre-trained StyleGAN generator $G$.

\subsection{Model Architecture}
\label{sec:architecture}
\method contains a semantic encoder and a flow-based transformation module to learn disentangled variables.

\vspace{-5pt}
\paragraph*{Flow-based transformation module.}

Conditioned on the given facial attributes of certain input face, the flow-based transformation module is to disentangle the semantic-irrelevant variables from the latent code. In this paper, we employ the conditional continuous normalizing flows~(CNFs)~\cite{kobyzev2020normalizing} as the flow-based transformation module for disentangled internal representations inspired by~\cite{abdal2021styleflow}. The conditional CNFs are optimized by neural ODE~\cite{chen2018neural}, whose mathematical basis in differential equations can be expressed as 
\begin{equation}
    \frac{\mathrm{d}\vct{z}}{\mathrm{d}t}=\Phi_\theta(\vct{z}(t),\vct{s},t),
\end{equation}
where $\vct{z}$ is the variable of Gaussian distribution, $t$ is the time variable, $\vct{s}$ is the semantic variables, and $\Phi_\theta$ is a neural network to produce $\frac{\mathrm{d}\vct{z}}{\mathrm{d}t}$, same as~\cite{abdal2021styleflow}. Therefore, the inverse inference of CNFs can be defined as 
\begin{equation}
    \vct{z}(t_1)=\vct{z}(t_0)+\int_{t_0}^{t_1}\Phi_\theta(\vct{z}(t),\vct{s},t)\mathrm{d}t,
    \label{eq:cnf_inverse}
\end{equation}
where $t_0$ and $t_1$ are the predefined start and end time. Here, $\vct{z}(t_0)=\vct{w}$ and $\vct{z}(t_1)\in \mathcal{N}(0, 1)$ should be irrelevant to $\vct{s}$. Therefore, the loss function to optimize the flow-based transformation module can be written as
\begin{equation}
    \mathcal{L}_{\mathrm{nll}} = - \log p(\vct{z}(t_0))+\int_{t_0}^{t_1}\operatorname{Tr}\left(\frac{\partial{\Phi}}{\partial{\vct{z}(t)}}\right)\mathrm{d}t,
    \label{equ:loss_nll}
\end{equation}
which maximizes the likelihood of the latent code $\vct{w}$. Face editing can be achieved by only changing semantic variables $\vct{s}$ in the reverse order of Eq.~\ref{eq:cnf_inverse}:
\begin{equation}
    \vct{w}_e=\vct{z}(t_1)-\int_{t_0}^{t_1}\Phi_\theta(\vct{z}(t),\vct{\hat{s}},t)\mathrm{d}t,
    \label{eq:forward}
\end{equation}
where $\vct{\hat{s}}$ is the target semantic variables. In practice, $\vct{s}$ can be set to the predictions of a pre-trained attribute classifier. However, this strategy does not provide well-disentangled $\vct{z}$ and $\vct{s}$; manipulating $\vct{s}$ would still change other undesired attributes.

\paragraph*{Semantic Estimator.}

\method opts to employ an additional semantic encoder $E_{\mathrm{s}}$ to estimate the semantic variables, which are jointly trained with the flow-based transformation module. As a result, $E_{\mathrm{s}}$ can adaptively adjust the semantic variables to produce semantic-irrelevant variables for the flows. Besides, $E_{\mathrm{s}}$ works directly from input faces instead of latent codes $\vct{w}$ which tends to be entangled~\cite{shen2020interface}.
In detail, $E_{\mathrm{s}}$ contains ResNet-34 \cite{he2016resnet} as the backbone, following three linear layers and a sigmoid activation function.
To inject the supervision signals of human annotations into the semantic variables, a pre-trained attribute classifier is distilled to $E_{\mathrm{s}}$:
\begin{equation}
    \mathcal{L}_{\mathrm{kd}} = \frac{1}{N}\sum_{i=1}^{N}\| C(\vct{I}) - E_{\mathrm{s}}(\vct{I})\|_2^2,
    \label{equ:loss_a}
\end{equation}
where $\|\cdot\|_2^2$ denotes Euclidean distance between predictions. 

\vspace{-5pt}
\subsection{Training and Inference}
\label{sec:trainandinfer}

\paragraph*{Disentangled learning.}
It is hard to make different variables disentangled well if simply jointly training two components.
To this end, we utilize the disentangled learning strategy in~\cite{chen2016infogan} to better regularize the semantic encoder and flow-based transformation module.
Specifically, the semantic variables $\vct{s}$ are randomly manipulated to $\vct{\hat{s}}$ sampling from $[0, 1]$, and then the faces are generated according to Eq.~\ref{eq:forward}. If $\vct{s}$ are disentangled well, $E_{\mathrm{s}}$ should be able to reconstrcut $\vct{\hat{s}}$. Therefore, we can maximize the mutual information between variables through:
\begin{equation}
\mathcal{L}_{\mathrm{mi}} = \frac{1}{N}\sum_{i=1}^{N}\| E_{\mathrm{s}}(\vct{\hat{I}}) - E_{\mathrm{s}}(\vct{I})\|_2^2,
\label{equ:loss_mi}
\end{equation}
where $\vct{\hat{I}}=G(\vct{w}_e)$ is the edited faces.

\paragraph*{Training and inference.}
To compress the unnecessary changes in the edited latent codes, we restrict the Euclidean distance between original and edited latent codes~\cite{yao2021latentTransformer}: 
\begin{equation}
    \mathcal{L}_{\mathrm{reg}}=\|\vct{w}-\vct{w}_e\|^2_2.
    \label{equ:loss_reg}
\end{equation}
Combining the Eqs.~\ref{equ:loss_nll},~\ref{equ:loss_a},~\ref{equ:loss_mi},~\ref{equ:loss_reg}, the overall training objective is to minimize the sum of all losses:
\begin{equation}
\mathcal{L}=\mathcal{L}_{\mathrm{nll}}+\mathcal{L}_{\mathrm{kd}}+\mathcal{L}_{\mathrm{mi}}+\mathcal{L}_{\mathrm{reg}}.
\end{equation}
During inference, we only need to change the desired semantic variables while keeping $\vct{z}$ and unrelated variables.

\section{Experiments}
\subsection{Implementation Details}
We performed \method with the official StyleGAN2 generator~\cite{karras2020stylegan2} pre-trained on FFHQ~\cite{karras2019stylegan}. We employed e4e~\cite{tov2021e4e} to invert the input faces, and trained the attribute classifiers on CelebA dataset~\cite{liu2015celeba}: a ResNet-34~\cite{he2016resnet} for knowledge distillation and a ResNet-50~\cite{he2016resnet} for evaluation. The flow-based transformation module follows the same architecture in~\cite{abdal2021styleflow} with $t_0$ and $t_1$ fixed as 0 and 1.
The first 69k faces of FFHQ are used for training while the rest and CelebA-HQ~\cite{karras2017progressive} are used for testing data~\cite{huang2023adaptive}.
All modules are jointly trained for 10,000 iterations with batch size 16, Adam optimizer~\cite{kingma2014adam}, learning rate $10^{-4}$, $\beta_1=0.9$, and $\beta_2=0.99$. All experiments are conducted on a single NVIDIA 3090 GPU.%

\vspace{-5pt}
\subsection{Experimental Results}

\begin{figure}[ht]
    \centering
    \includegraphics[width=\linewidth]{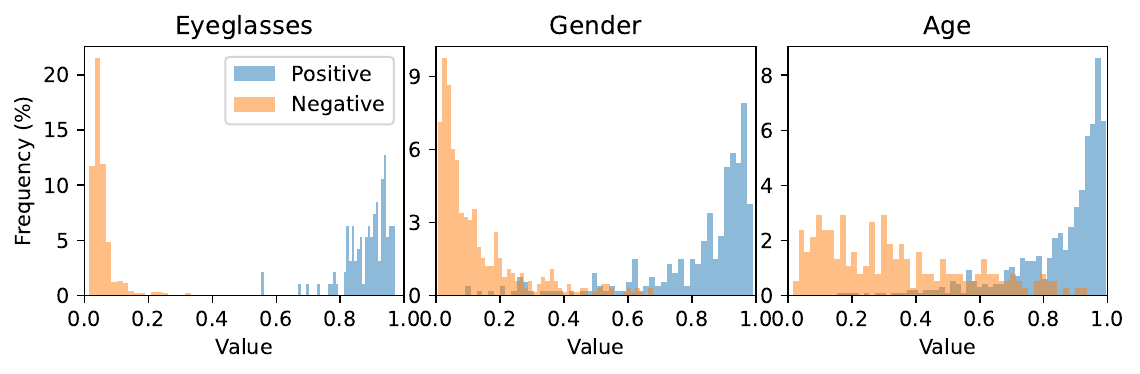}
    \vspace{-20pt}
    \caption{Histgram of semantic variables predicted by $E_{s}$.}
    \label{fig:distribution}
\end{figure}

\paragraph*{Visualization of semantic variables.}
Fig.~\ref{fig:distribution} visualizes the semantic variables of three facial attributes, \ie, \textit{Eyeglasses}, \textit{Gender}, and \textit{Age}, predicted by $E_{s}$. The results show that $E_{s}$ can learn meaningful values for different variables. In terms of \textit{Eyeglasses}, there is a clear decision boundary since it is distinct to identify the faces with/without glasses. On the contrary, \textit{Age} exhibits overlapping, which conforms to the practical scenario of facial aging process. 
In summary, Fig.~\ref{fig:distribution} validates that binary attributes are insufficient to effectively describe the strength of attributes, emphasizing the importance of incorporating our proposed semantic decomposition.

\vspace{-10pt}

\begin{figure*}[t]
	\centering
	\subfloat[Editing single attributes.]{
            \label{fig:quantitative_results_single_attributes}
	       \includegraphics[width=\columnwidth]{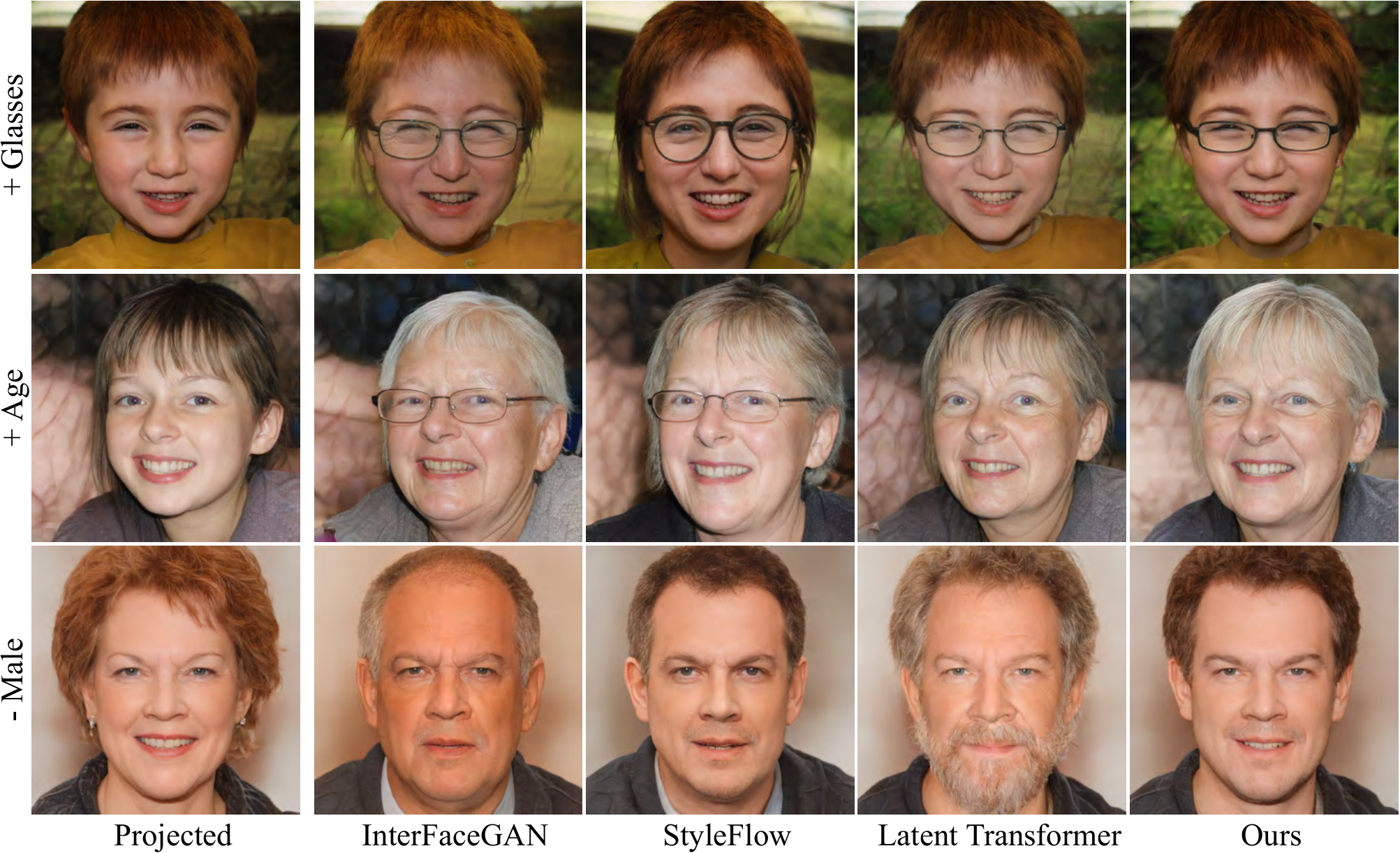}
		}
	\subfloat[Editing multiple attributes.]{
        \label{fig:quantitative_results_multi_attributes}
		\includegraphics[width=\columnwidth]{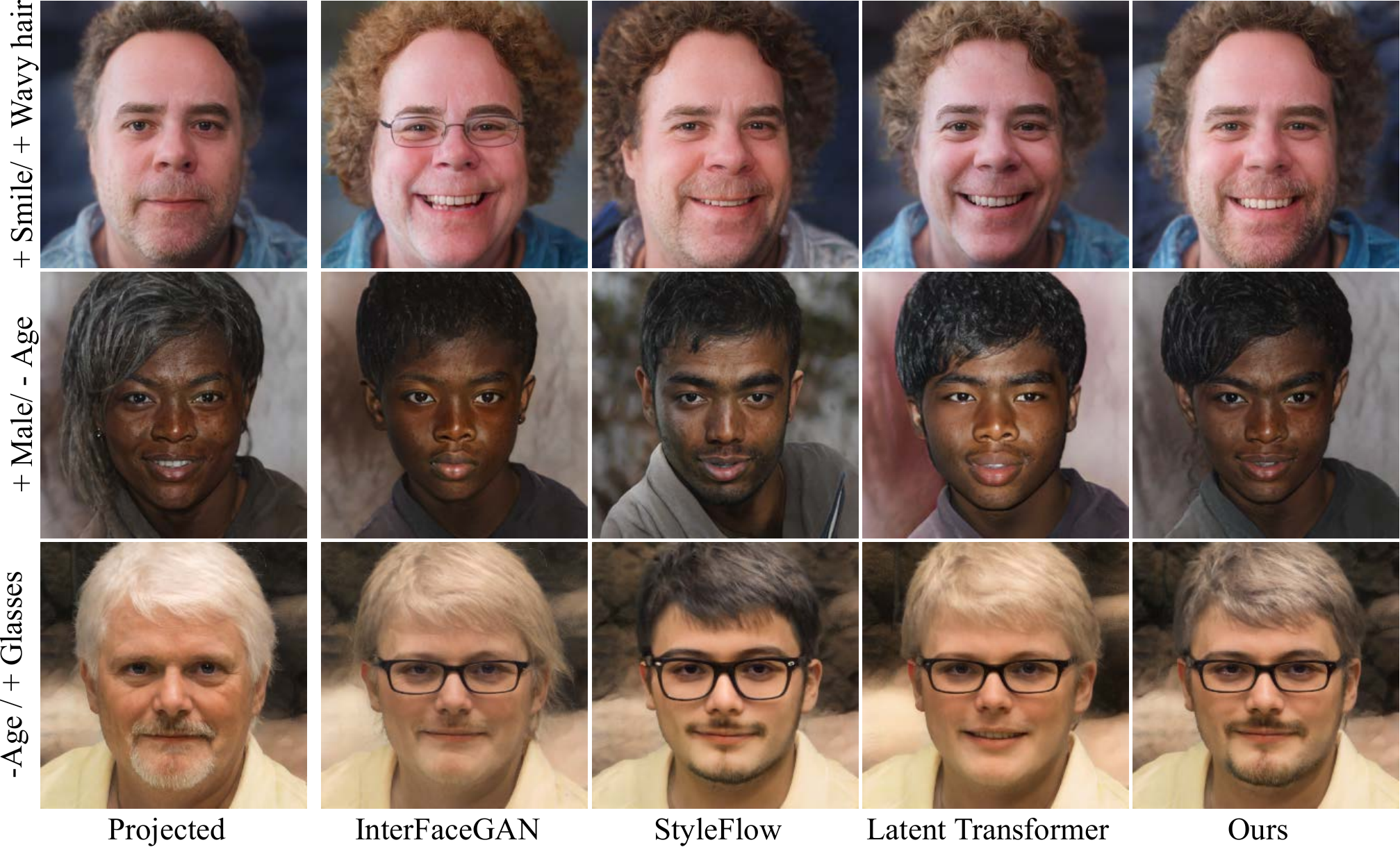}
		}
        \vspace{-8pt}
	\caption{Qualitative comparisons with recent methods for face editing. The competitors produce unexpected changes in unrelated attributes or fail to handle the hard cases when editing single or multiple attributes.
    }
        \label{fig:quantitative_comparisons}
\end{figure*}

\noindent\textbf{Qualitative evaluation.} 
Fig.~\ref{fig:quantitative_comparisons} showcases example results on manipulating the faces into target attributes.

We compare our method with InterfaceGAN~\cite{shen2020interface}, Latent Transformer~\cite{yao2021latentTransformer} and StyleFlow~\cite{abdal2021styleflow} and manually select the best editing strength.
Previous methods~\cite{abdal2021styleflow,shen2020interface,yao2021latentTransformer} usually change one's identity or other unrelated attributes, such as adding glasses when aging. 
Fig.~\ref{fig:quantitative_results_single_attributes} showcases the effectiveness of our \method in successfully disentangling highly correlated attributes~\cite{shen2020interface}. Fig.~\ref{fig:quantitative_results_multi_attributes} shows that \method achieves the best results in multi-attribute editing. Our proposed \method excels in producing photo-realistic and disentangled manipulations while preserving the identity and unrelated attributes.

\vspace{-10pt}
\begin{figure}[!ht]
    \centering
    \includegraphics[width=\linewidth]{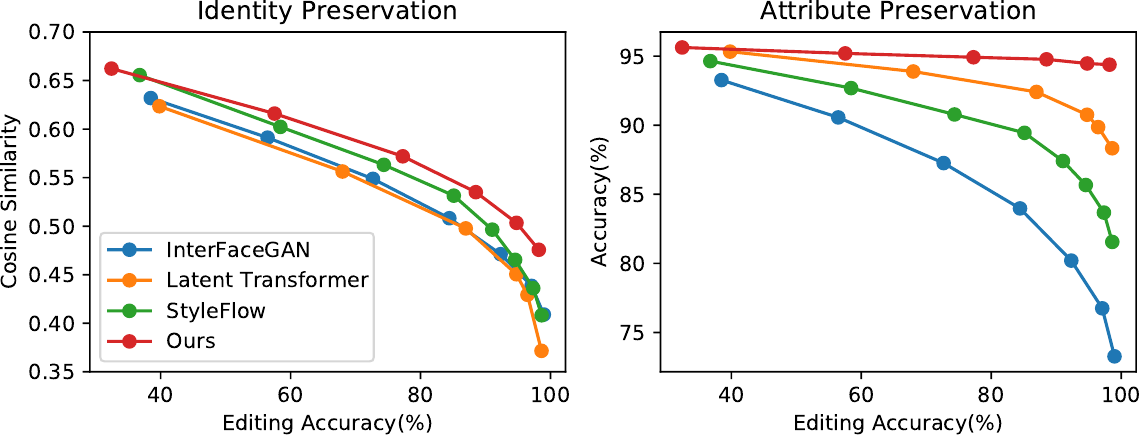}
    \vspace{-20pt}
    \caption{Quantitative results for multi-attribute editing. A higher curve indicates better performance.
    }
    \label{fig:accuracy}
\end{figure}

\paragraph*{Quantitative evaluation.}
We desire a higher proportion of successfully manipulated samples with fewer changes in identity and unrelated attributes. We use three widely used metrics to compare different methods quantitatively, including editing accuracy, attribute preservation accuracy, and identity preservation~\cite{huang2023adaptive}. An attribute classifier with ResNet-50 is trained from scratch to predict the attributes of the manipulated faces, which can be used to measure editing accuracy and attribute preservation. Identity preservation is the cosine similarity between the facial embeddings extracted by~\cite{deng2019arcface}.
We gradually increase the editing strength of manipulation towards the opposite attribute until the editing accuracy reaches 99\%. Consequently, we can draw the identity/attribute preservation curves \textit{w.r.t} editing accuracy. The qualitative results are shown in Fig.~\ref{fig:accuracy}. Our method outperforms three competitors by a large margin, indicating better disentangled and accurate face editing.

\vspace{-5pt}
\begin{figure}[h]
    \centering
    \includegraphics[width=\linewidth]{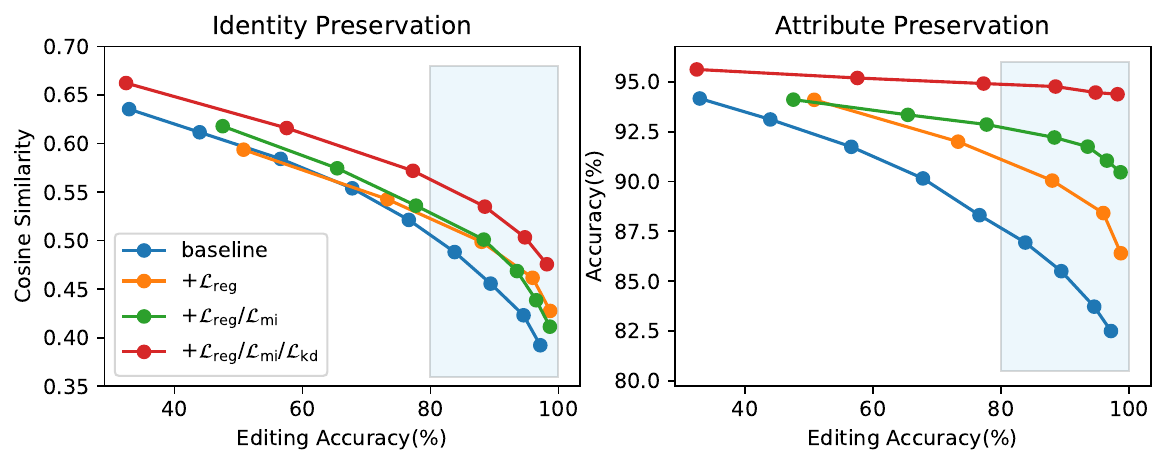}
    \vspace{-20pt}
    \caption{Quantitative ablation study of proposed components.}
    \label{fig:ablation}
\end{figure}
\vspace{-15pt}

\subsection{Ablation Study}

\begin{figure}[t]
    \centering
    \includegraphics[width=1.0\linewidth]{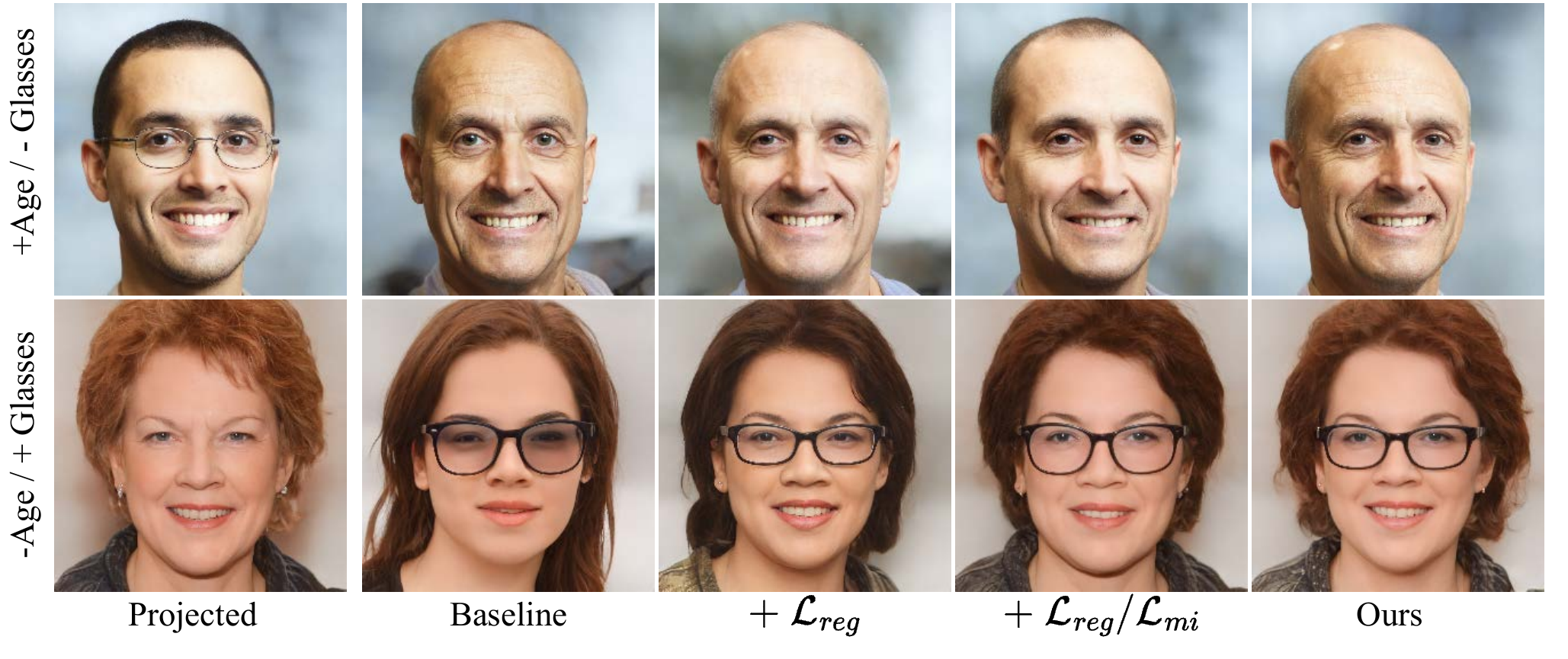}
    \vspace{-20pt}
    \caption{Qualitative ablation study of proposed components.}
    \label{fig:ablation_showcase}
\end{figure}

We conduct ablation studies to validate the effectiveness of different components in our methods. We start from the baseline StyleFlow~\cite{abdal2021styleflow}, which is directly optimized by $\mathcal{L}_{\mathrm{nll}}$. Then we gradually add $\mathcal{L}_{\mathrm{reg}}$, $\mathcal{L}_{\mathrm{mi}}$, and $\mathcal{L}_{\mathrm{kd}}$. It is worthwhile to note that a pre-trained attribute classifier is employed as the semantic encoder when adding $\mathcal{L}_{\mathrm{mi}}$, and the semantic encoder is jointly trained when adding $\mathcal{L}_{\mathrm{kd}}$. The quantitative and qualitative results are presented in Fig.~\ref{fig:ablation} and Fig.~\ref{fig:ablation_showcase}, respectively.
Interestingly, adding $\mathcal{L}_{\mathrm{reg}}$ can achieve significant improvements over baseline, indicating that only inverse inference during optimization cannot preserve identity/attributes. Although $\mathcal{L}_{\mathrm{mi}}$ with a pre-trained classifier is helpful for attribute preservation, the identity preservation unexpectedly drops. We argue that the pre-trained classifier would over-manipulate the latent codes to unrelated attributes while ignoring identity. On the contrary, all three components~(our \method) can address this issue as the semantic encoder is involved in the training process to estimate better semantic variables.

\vspace{-5pt}

\section{Conclusions}
In this paper, we introduce \method for face editing with a semantic decomposition in original latent space using continuous conditional normalizing flows. A semantic encoder is introduced to estimate the proper semantic variables. The flow-based transformation module enables nonlinear editing and produces semantic-irrelevant variables, under the conditions of semantic variables. A disentangled learning strategy is adopted to eliminate the entanglement among the attributes. Extensive experiments demonstrate that our \method can generate photo-realistic results with better disentanglement.


\newpage
\bibliographystyle{IEEEbib}
\bibliography{refs}

\begin{thebibliography}{10}

\bibitem{karras2019stylegan}
Tero Karras, Samuli Laine, and Timo Aila,
\newblock ``A style-based generator architecture for generative adversarial
  networks,''
\newblock in {\em CVPR}, 2019.

\bibitem{karras2020stylegan2}
Tero Karras, Samuli Laine, Miika Aittala, Janne Hellsten, Jaakko Lehtinen, and
  Timo Aila,
\newblock ``Analyzing and improving the image quality of stylegan,''
\newblock in {\em CVPR}, 2020.

\bibitem{goodfellow2014generative}
Ian Goodfellow, Jean Pouget-Abadie, Mehdi Mirza, Bing Xu, David Warde-Farley,
  Sherjil Ozair, Aaron Courville, and Yoshua Bengio,
\newblock ``Generative adversarial nets,''
\newblock in {\em NeurlIPS}, 2014.

\bibitem{richardson2021psp}
Elad Richardson, Yuval Alaluf, Or~Patashnik, Yotam Nitzan, Yaniv Azar, Stav
  Shapiro, and Daniel Cohen-Or,
\newblock ``Encoding in style: A stylegan encoder for image-to-image
  translation,''
\newblock in {\em CVPR}, 2021.

\bibitem{tov2021e4e}
Omer Tov, Yuval Alaluf, Yotam Nitzan, Or~Patashnik, and Daniel Cohen-Or,
\newblock ``Designing an encoder for stylegan image manipulation,''
\newblock {\em TOG}, vol. 40, no. 4, pp. 1--14, 2021.

\bibitem{abdal2022clip2stylegan}
Rameen Abdal, Peihao Zhu, John Femiani, Niloy Mitra, and Peter Wonka,
\newblock ``{CLIP}2{StyleGAN}: Unsupervised extraction of stylegan edit
  directions,''
\newblock in {\em SIGGRAPH}, 2022.

\bibitem{abdal2021styleflow}
Rameen Abdal, Peihao Zhu, Niloy~J Mitra, and Peter Wonka,
\newblock ``Styleflow: Attribute-conditioned exploration of stylegan-generated
  images using conditional continuous normalizing flows,''
\newblock {\em TOG}, vol. 40, no. 3, pp. 1--21, 2021.

\bibitem{shen2020interface}
Yujun Shen, Jinjin Gu, Xiaoou Tang, and Bolei Zhou,
\newblock ``Interpreting the latent space of gans for semantic face editing,''
\newblock in {\em CVPR}, 2020.

\bibitem{yao2021latentTransformer}
Xu~Yao, Alasdair Newson, Yann Gousseau, and Pierre Hellier,
\newblock ``A latent transformer for disentangled face editing in images and
  videos,''
\newblock in {\em ICCV}, 2021.

\bibitem{yang2021discovering}
Huiting Yang, Liangyu Chai, Qiang Wen, Shuang Zhao, Zixun Sun, and Shengfeng
  He,
\newblock ``Discovering interpretable latent space directions of {GANs} beyond
  binary attributes,''
\newblock in {\em CVPR}, 2021.

\bibitem{wu2021stylespace}
Zongze Wu, Dani Lischinski, and Eli Shechtman,
\newblock ``Stylespace analysis: Disentangled controls for stylegan image
  generation,''
\newblock in {\em CVPR}, 2021.

\bibitem{yuan2023dofam}
Yifan Yuan, Siteng Ma, Hongming Shan, and Junping Zhang,
\newblock ``{DO-FAM}: Disentangled non-linear latent navigation for facial
  attribute manipulation,''
\newblock in {\em ICASSP}, 2023.

\bibitem{huang2023adaptive}
Zhizhong Huang, Siteng Ma, Junping Zhang, and Hongming Shan,
\newblock ``Adaptive nonlinear latent transformation for conditional face
  editing,''
\newblock in {\em ICCV}, 2023.

\bibitem{harkonen2020ganspace}
Erik H{\"a}rk{\"o}nen, Aaron Hertzmann, Jaakko Lehtinen, and Sylvain Paris,
\newblock ``Ganspace: Discovering interpretable {GAN} controls,''
\newblock in {\em NeurlIPS}, 2020.

\bibitem{shen2021closed}
Yujun Shen and Bolei Zhou,
\newblock ``Closed-form factorization of latent semantics in {GANs},''
\newblock in {\em CVPR}, 2021.

\bibitem{radford2021clip}
Alec Radford, Jong~Wook Kim, Chris Hallacy, Aditya Ramesh, Gabriel Goh,
  Sandhini Agarwal, Girish Sastry, Amanda Askell, Pamela Mishkin, Jack Clark,
  et~al.,
\newblock ``Learning transferable visual models from natural language
  supervision,''
\newblock in {\em ICML}, 2021.

\bibitem{patashnik2021styleclip}
Or~Patashnik, Zongze Wu, Eli Shechtman, Daniel Cohen-Or, and Dani Lischinski,
\newblock ``Style{CLIP}: Text-driven manipulation of stylegan imagery,''
\newblock in {\em ICCV}, 2021.

\bibitem{chen2016infogan}
Xi~Chen, Yan Duan, Rein Houthooft, John Schulman, Ilya Sutskever, and Pieter
  Abbeel,
\newblock ``Info{GAN}: Interpretable representation learning by information
  maximizing generative adversarial nets,''
\newblock in {\em NeurlIPS}, 2016.

\bibitem{kobyzev2020normalizing}
Ivan Kobyzev, Simon~JD Prince, and Marcus~A Brubaker,
\newblock ``Normalizing flows: An introduction and review of current methods,''
\newblock {\em TPAMI}, vol. 43, no. 11, pp. 3964--3979, 2021.

\bibitem{chen2018neural}
Ricky~TQ Chen, Yulia Rubanova, Jesse Bettencourt, and David~K Duvenaud,
\newblock ``Neural ordinary differential equations,''
\newblock in {\em NeurlIPS}, 2018.

\bibitem{he2016resnet}
Kaiming He, Xiangyu Zhang, Shaoqing Ren, and Jian Sun,
\newblock ``Deep residual learning for image recognition,''
\newblock in {\em CVPR}, 2016.

\bibitem{liu2015celeba}
Ziwei Liu, Ping Luo, Xiaogang Wang, and Xiaoou Tang,
\newblock ``Deep learning face attributes in the wild,''
\newblock in {\em ICCV}, 2015.

\bibitem{karras2017progressive}
Tero Karras, Timo Aila, Samuli Laine, and Jaakko Lehtinen,
\newblock ``Progressive growing of {GAN}s for improved quality, stability, and
  variation,''
\newblock in {\em arXiv preprint arXiv:1710.10196}, 2017.

\bibitem{kingma2014adam}
Diederik~P Kingma and Jimmy Ba,
\newblock ``Adam: A method for stochastic optimization,''
\newblock in {\em ICLR}, 2015.

\bibitem{deng2019arcface}
Jiankang Deng, Jia Guo, Niannan Xue, and Stefanos Zafeiriou,
\newblock ``Arcface: Additive angular margin loss for deep face recognition,''
\newblock in {\em CVPR}, 2019.

\end{thebibliography}

\end{document}